\documentclass[sigconf]{acmart}

\AtBeginDocument{%
  }

\copyrightyear{2024}
\acmYear{2024}
\setcopyright{rightsretained}
\acmConference[CIKM '24]{Proceedings of the 33rd ACM International Conference on Information and Knowledge Management}{October 21--25, 2024}{Boise, ID, USA}
\acmBooktitle{Proceedings of the 33rd ACM International Conference on Information and Knowledge Management (CIKM '24), October 21--25, 2024, Boise, ID, USA}
\acmDOI{10.1145/3627673.3679797}
\acmISBN{979-8-4007-0436-9/24/10}

\usepackage[ruled, vlined, linesnumbered]
{algorithm2e}

\usepackage{multirow}
\usepackage{graphicx}
\usepackage{pifont}
\usepackage{caption}
\usepackage{titlesec}
\newcommand{\cmark}{\ding{51}}%
\newcommand{\xmark}{\ding{55}}%

\titlespacing{\section}{0pt}{*0.5}{*0.5}
\titlespacing{\subsection}{0pt}{*0.5}{*0.5}
\titlespacing{\subsubsection}{0pt}{*0.5}{*0.5}

\begin{document}

\title{HiQuE: Hierarchical Question Embedding Network for Multimodal Depression Detection}

\author{Juho Jung}
\affiliation{%
    \department{Dept. of Applied Artificial Intelligence}
  \institution{Sungkyunkwan University}
  \city{Seoul}
  \country{South Korea}
  }
\email{jhjeon9@g.skku.edu}

\author{Chaewon Kang}
\affiliation{%
    \department{Dept. of Applied Artificial Intelligence}
  \institution{Sungkyunkwan University}
  \city{Seoul}
  \country{South Korea}
  }
\email{codnjs3@g.skku.edu}

\author{Jeewoo Yoon}
\affiliation{%
  \institution{Raondata}
  \city{Seoul}
  \country{South Korea}
  }
\email{jeewooyoon@raondata.ai}

\author{Seungbae Kim}
\affiliation{%
\department{Computer Science and Engineering}
  \institution{University of South Florida}
  \city{Tampa, Florida}
  \country{USA}
  }
\email{seungbae@usf.edu}

\author{Jinyoung Han}
\authornote{Corresponding Author.}
\affiliation{%
    \department{Dept. of Applied Artificial Intelligence}
  \institution{Sungkyunkwan University}
  \city{Seoul}
  \country{South Korea}
  }
\email{jinyounghan@skku.edu}



\begin{abstract}
  The utilization of automated depression detection significantly enhances early intervention for individuals experiencing depression. Despite numerous proposals on automated depression detection using recorded clinical interview videos, limited attention has been paid to considering the hierarchical structure of the interview questions. In clinical interviews for diagnosing depression, clinicians use a structured questionnaire that includes routine baseline questions and follow-up questions to assess the interviewee's condition. This paper introduces \textbf{HiQuE} (\textbf{Hi}erarchical \textbf{Qu}estion \textbf{E}mbedding network), a novel depression detection framework that leverages the hierarchical relationship between primary and follow-up questions in clinical interviews. HiQuE can effectively capture the importance of each question in diagnosing depression by learning mutual information across multiple modalities. We conduct extensive experiments on the widely-used clinical interview data, DAIC-WOZ, where our model outperforms other state-of-the-art multimodal depression detection models and emotion recognition models, showcasing its clinical utility in depression detection.
\end{abstract}



\maketitle

\section{Introduction}
The diagnosis of depression in clinical settings often involves the use of interview-based instruments~\cite{smith2013diagnosis}, in which mental health experts conduct clinical interviews with patients, assessing their symptoms~\cite{smith2013diagnosis, paykel1985clinical, stuart2014comparison,young1987research}. Due to the gradual and varied manifestation of depressive symptoms among individuals~\cite{cummins2013diagnosis, dibekliouglu2017dynamic, dibekliouglu2015multimodal, morales2016speech}, clinicians employ a structured interview process, which involves specific questionnaires and criteria, to detect a wide range of verbal and non-verbal symptoms of depression in patients' speech, behavior, facial expressions, and immediate responses during conversations~\cite{white2022articulation,timbremont2004assessing,mojtabai2013clinician}.

Interview-based diagnostic methods have proven highly effective in interpreting patient responses~\cite{paykel1985clinical, guidi2010clinical}. Mental health experts strategically incorporate follow-up questions in conjunction with primary questions to gather additional information from patients, thereby enhancing their understanding of the exhibited depressive symptoms~\cite{stuart2014comparison}. In a case where a response from an initial follow-up question is insufficient for diagnosis, further follow-up questions can be employed to synthesize the patient’s responses. By employing hierarchical questions during clinical interviews, clinicians can obtain a comprehensive understanding of depressive signals and the patient's overall condition, leading to improved diagnostic accuracy~\cite{stuart2014comparison, dibekliouglu2017dynamic}.

\begin{figure}[t]
  \centering
      \includegraphics[width=7.5cm]{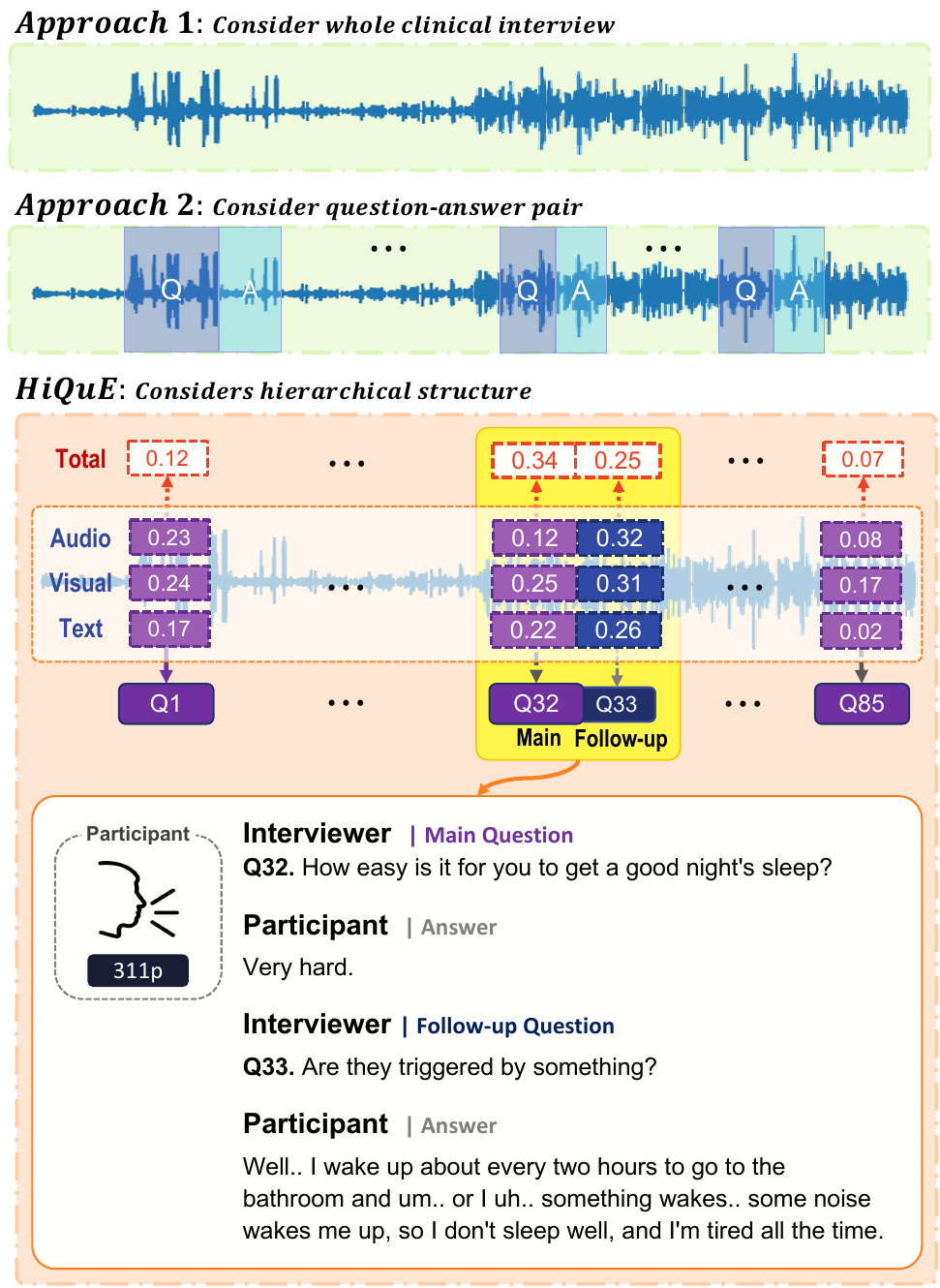}
  \caption{Previous research focused on learning the whole clinical interview sequences or their question-and-answer segments using a single modality. Our novel model, HiQuE, considers the hierarchy of questions incorporating multiple modalities to improve its effectiveness in depression detection. Numerical values in the boxes represent attention scores.}
  \label{fig:speech_production}
\end{figure}

Using clinical interview data, many scholars have proposed methods that can detect depression by analyzing revealed verbal (e.g., textual) or non-verbal (e.g., visual or acoustic) signals. Some studies have delved into visual cues, encompassing facial expressions and head poses~\cite{alghowinem2013head, alghowinem2016multimodal, guo2022automatic, nasser2020review}. Additionally, a series of investigations has focused on acoustic and textual cues, with the goal of diagnosing depression based on linguistic patterns, vocal qualities, pitch, and loudness~\cite{al2018detecting, ma2016depaudionet, zhang2021depa, sardari2022audio, yadav2021review}. Nevertheless, these approaches have treated the entire input sequence as a singular entity, disregarding the structured nature of clinical interviews. Given the use of structured questionnaires and conversational exchanges during these interviews, considering the interview data as a single input sequence can be less effective for depression detection in clinical settings~\cite{yin2019multi,guohou2020reveals}.
A few studies have sought to consider the conversations, including questions and answers during the clinical interviews~\cite{williamson2016detecting, chen2022speechformer, xezonaki2020affective, mallolragolta19_interspeech, niu2021hcag,yin2019multi}. However, their focus has primarily been on analyzing questions and answers in an interview without specifically modeling the relationship between primary and follow-up questions based on question types, which can be crucial in modeling and analyzing the structure of a clinical interview. Besides, there has been a lack of analysis on the interaction among multi-modalities in analyzing structured clinical interviews~\cite{rohanian2019detecting, lam2019context, shen2022automatic, al2018detecting, majumder2019dialoguernn, dong2021hierarchical,flores2021depression,yin2019multi}; different modalities can be different cues depending on questions and answer types. 

To address these challenges, we propose HiQuE (Hierarchical Question Embedding network), a novel depression detection framework that leverages the hierarchical relationship between primary and follow-up questions in a clinical interview. Inspired by the clinical interview strategy employed by medical professionals for diagnosing depression, HiQuE incorporates a hierarchical embedding structure and interview-specific attention modules. These modules enable HiQuE to comprehensively assess the mutual information between multiple modalities within interviews, replicating the diagnostic approach used by clinicians. As illustrated in Figure~\ref{fig:speech_production}, the interview sequence is divided into primary questions and their corresponding follow-up questions. Using the question-aware module, HiQuE calculates the significance of each question and effectively enhances the mutual information across modalities using cross-modal attention, resulting in accurate depression diagnosis. The contributions of this study can be summarized as follows:

\begin{itemize}
\item To the best of our knowledge, HiQuE is the first attempt that analyzes the significance of all questions posed by the interviewer by explicitly categorizing them as primary and follow-up questions, considering their order and relationship. Our publicly available code\footnote{\url{https://github.com/JuHo-Jung/HiQuE}} encompasses both the hierarchical question embedding process and the HiQuE.

\item This is the first interpretable multi-modal analysis conducted in a clinical interview context by analyzing both intra-modality and inter-modality attention scores. The quantitative evaluation of the interaction and importance of different modalities in depression detection provides deeper insights into the complex dynamics of clinical interviews.

\item HiQuE achieves the state-of-the-art performance on the DAIC-WOZ dataset, among other multimodal emotion recognition models as well as prior depression detection models that utilized the DAIC-WOZ dataset. HiQuE also demonstrates superior adaptation to the E-DAIC-WOZ dataset, highlighting its generalizability to unseen question scenarios.
\end{itemize}

\begin{figure*}[t]
  \centering
  \includegraphics[width=0.95\linewidth]{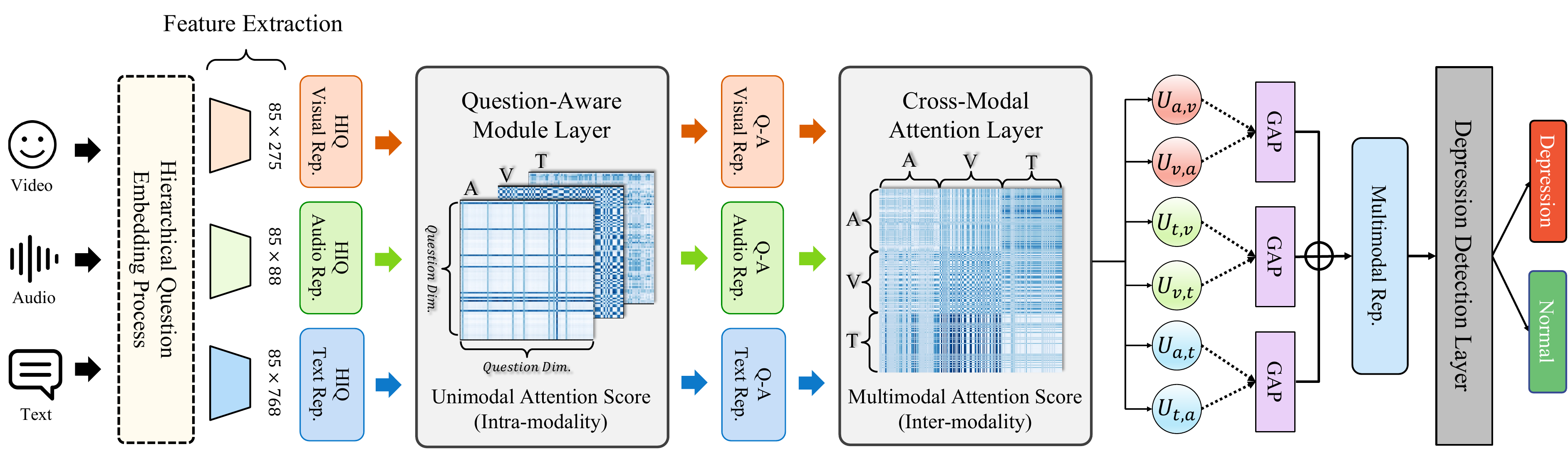}
  \caption{An overall architecture of the HiQuE's multimodal depression detection process, where \textit{HIQ} and \textit{Q-A} indicate \textit{Hierarchical Question Embedded} and \textit{Question-Aware, respectively}.}
  \label{fig:overall_architecture}
\end{figure*}

\section{Related Work}

\subsection{Multimodal Expressions of Major Depressive Disorder in Clinical Interviews}

Researchers have identified distinctive features across various modalities, including acoustic patterns, visual characteristics, and language usage in clinical interviews. For instance, individuals with depression often exhibit specific acoustic features, such as slower speaking rates, lower pitch ranges, and reduced loudness~\cite{white2022articulation, wang2019acoustic, ellgring1996vocal, vicsi2012examination, cannizzaro2004voice, kraepelin1921manic, alpert2001reflections} as well as visual features, including discernible facial expressions characterized by sadness, minimal head movement~\cite{alghowinem2013head}, unstable facial expressions~\cite{waxer1974nonverbal, cummins2015review}, and irregular eye-gazing patterns~\cite{li2016eye, gupta2014multimodal}. Moreover, they often reveal negative emotions in language, utilize a higher frequency of first-person pronouns, and exhibit intense focus on specific words~\cite{yang2012detecting, al2018absolute, rude2004language}. These findings highlight the crucial role of incorporating multiple modalities for effective depression detection to develop a comprehensive understanding of an individual's symptoms~\cite{dibekliouglu2017dynamic, dibekliouglu2015multimodal, morales2016speech}.

\subsection{Automatic Depression Detection}

There have been a considerable number of proposals to detect depression by developing machine learning algorithms or natural language processing techniques~\cite{lin2020towards,morales2017cross}. Initially, a substantial efforts were dedicated to extracting representative features~\cite{sun2017random, sturim2011automatic} and creating single-modality models for depression detection~\cite{gulati2020conformer, wang2020transformer,nasser2020review,yadav2021review,yadav2023novel}. Furthermore, as Multimodal Sentiment Analysis (MSA)~\cite{morency2011towards,soleymani2017survey} gained momentum with the recognition of various verbal and non-verbal symptoms of depression in psychological research, researchers made significant attempts to incorporate context-aware attention~\cite{chauhan2019context} and multimodal attention~\cite{ghosal2018contextual} to capture diverse information across multiple modalities~\cite{lam2019context,saggu2022depressnet,fang2023multimodal,ye2021multi,yoon2022d,dibekliouglu2015multimodal}. Recently, there have been attempts that analyze the word-sentence relations on interviewee's answers~\cite{xezonaki2020affective,zogan2023hierarchical,yin2019multi,mallolragolta19_interspeech,rohanian2019detecting} as well as the correlation between question-answer pairs~\cite{niu2021hcag,flores2022transfer,williamson2016detecting}, which can be cued in identifying depression. Unfortunately, no research exists yet that explains how an attention score of modality manifests in a specific question or how the degree of modality reflection changes with the sequence of primary and follow-up questions and answers, which can be crucial in modeling and analyzing the structure of a clinical interview. To bridge this gap, we introduce the first interpretable multimodal depression detection framework that leverages the hierarchical relationship between primary and follow-up questions in a clinical interview.

\section{Clinical Interview Dataset}
To train our proposed method for the depression detection task, we use the DAIC-WOZ dataset~\cite{gratch2014distress}, which is a subset of the widely used dataset called Distress Analysis Interview Corpus (DAIC)~\cite{valstar2016avec}. The DAIC-WOZ dataset comprises clinical interviews conducted to diagnose psychological distress disorders. These interviews involve Wizard-of-Oz interactions, where an AI virtual interviewer named Ellie is controlled by a human interviewer located remotely. The dataset consists of speech samples from 189 participants, including audio/visual features, raw audio files, and interview transcripts. Following the prescribed guidelines, we split the dataset into 107 training samples, 35 validation samples, and 47 test samples. 


\subsection{Data Augmentation with Random Sampling}
\label{augmentation}
The DAIC-WOZ dataset suffers from a significant class imbalance, with a higher proportion of non-depression samples.
Some prior studies addressed this issue by employing data augmentation techniques like random masking~\cite{bailey2021gender, ma2016depaudionet,shen2022automatic}. Inspired by these, we tripled the size of the depression dataset by randomly masking 10 out of 85 questions in each $85 \times N$ question-embedded interview sequence, aligning it with the size of the non-depression dataset during training. Specifically, we first segmented the interview sequences into question-answer (Q-A) pairs based on timestamps, starting from the interviewer's question to the participant's response. Then, we randomly masked ten Q-A pairs per interview, corresponding to the interviewer's questions. Unused questions were replaced with zero vectors. This augmentation process was applied during only training.



\section{Hierarchical Question Embedding Network}
    \subsection{Problem Statement}
        Suppose we have a set of depression dataset $\textit{C}=\{c_i\}_{i=1}^{|C|}$, where $c_i$ contains the multi-modal inputs including audio, video, and text sequences; $X_a \in \mathbb{R}^{L_a \times d_a}$, $X_v \in \mathbb{R}^{L_v \times d_v}$, and $X_t \in \mathbb{R}^{L_t \times d_t}$, where $L_m$ represents the sequence length and $d$ indicates the feature dimension. Given the hierarchical structure of interview questions, we segment the interview sequence into question-answer pairs. Specifically, each input sequence is defined as $S = \{ s_i \}_{i=1}^{n}$, where $s_i = (question_i, answer_i)$. We then annotate the input sequence $S$ with corresponding hierarchical positions in the hierarchical question embedding process, denoted as $\hat{S} = \{ (s_i, pos_i) \}_{i=1}^{n}$, where $pos_i$ indicates the hierarchical position of the $question_i$. Finally, the proposed model predicts an individual $c_i$ depression symptom $\hat{y} \in \{normal, depression\}$.

    \subsection{Overall Architecture}
        The proposed method, HiQuE, as shown in Figure~\ref{fig:overall_architecture}, consists of three layers: (i) \emph{Question-Aware Module}, (ii) \emph{Cross-Modal Attention}, and (iii) \emph{Depression Detection}. HiQuE categorizes interview sequences into main and follow-up questions using a hierarchical question embedding process. Audio, visual, and text features are extracted separately, and the \emph{Question-Aware Module} generates Question-Aware representations for each feature. These are combined in the \emph{Cross-Modal Attention} layer to create a final multimodal representation, which the \emph{Depression Detection} layer uses to predict the presence of depression.

    \begin{figure}[t]
      \centering
      \includegraphics[width=\linewidth]{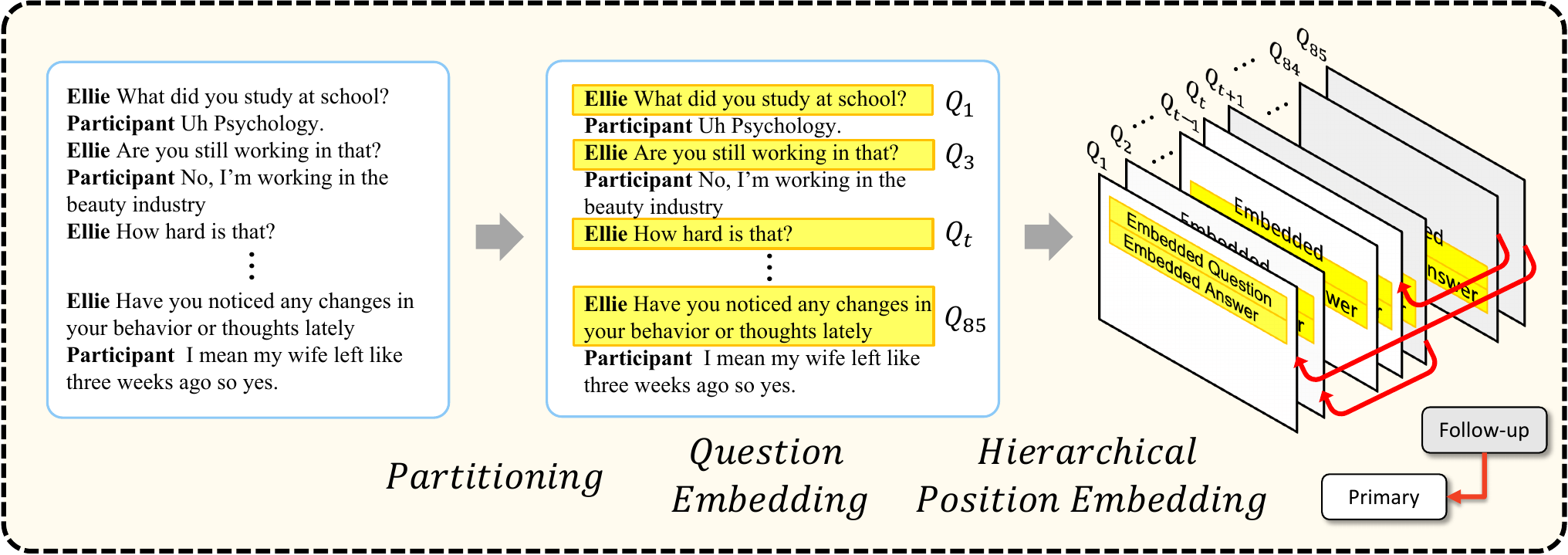}
      \caption{Hierarchical Question Embedding Process.}
      \label{fig:hierachical_qe}
      \vspace{-0.1in}
    \end{figure}

    \subsection{Hierarchical Question Embedding Process}
        As depicted in Figure~\ref{fig:hierachical_qe}, interviewer Ellie's questions are categorized into 85 topics based on content, following Gong et al.'s approach~\cite{gong2017topic}. Each question is associated with a specific topic code, such as labeling ``\textit{How has seeing a therapist affected you?}'' as \textit{therapist\_affect} and ``\textit{Where are you from originally?}'' as \textit{origin}. These questions are further categorized into 66 primary and 19 follow-up questions based on content and order. For a complete list of the questions, please refer to Table~\ref{tab:questionlist}.        
        Finally, we systematically tag each question based on its hierarchical order, specifically when a follow-up question follows a primary question or when a follow-up question follows a previous follow-up question. For instance, where the question sequence is ``\textit{What did you study at school?}'', ``\textit{Are you still working in that?}'', and ``\textit{How hard is that?}'', the hierarchical order would be primary -- follow-up -- follow-up.

        An overall process of hierarchical question embedding is depicted in Figure~\ref{fig:hierachical_qe}. Interview sequences are represented as unimodal raw sequences $X_{m}$, where $m$ denotes $modality \in \left\{a, v, t\right\}$, respectively. Sequences $X_m$ are partitioned into segments $ S=\left\{ s_i \right\}_{i=1}^{n}$ based on question and answer boundaries. Note that the number of segments, $n$, may vary for each sample due to differences in the type and number of questions employed during each interview. Then, each segment is split into a question segment and an answer segment; $ S=\left\{ (q_1, a_1)  \cdots (q_i, a_i) \cdots (q_{85}, a_{85}) \right\}$. 
        After partitioning, segments are labeled with topic codes corresponding to each question and given hierarchical position embeddings based on their relationships. Specifically, as shown in Figure~\ref{fig:hierachical_qe}, we assign the previous question's \textit{Topic id} to the follow-up question. These hierarchical positions are incorporated into the representation before feeding them into the model. This embedding process ensures uniform vector shapes by replacing unused questions with zero vectors, resulting in 85-dimensional representations for all samples. 

    \subsection{Feature Extraction}

        \subsubsection{Audio Feature: }
        For audio feature extraction, we utilize the open-Source Media Interpretation by Large feature-space Extraction (openSMILE)~\cite{eyben2010opensmile}, along with the extended Geneva Minimalistic Acoustic Parameter Set~\cite{eyben2015geneva}. These features encompass 88 functionals, including loudness, MFCCs, and other characteristics that aid in discerning emotions in speech. Consequently, each interviewee's audio features are represented as 85 $\times$ 88-dimensional vectors, where 85 denotes the question embedding dimension. These audio features are then processed using a transformer encoder.

        \subsubsection{Visual Feature: }
        Due to privacy concerns, the dataset only offers visual features extracted via the Constrained Local Neural Fields (CLNF) algorithm~\cite{baltrusaitis2013constrained}, a widely-used approach for facial landmark localization and face recognition. To address the variation in interview duration for each answer, we first extract 68 facial landmarks from each frame (at a rate of 1 frame per second) within each segment, considering their respective $x$ and $y$ coordinates. We then compute mean and variance vectors within each segment and concatenate the $x$ and $y$ coordinates. This results in 85 $\times$ 272-dimensional vectors per participant, with zero vectors used for segments where a face is not detected.

        \subsubsection{Text Feature: }
        For text feature extraction, we segment the interview transcripts into sections corresponding to individual answers for each question. We next leverage the pre-trained RoBERTa~\cite{liu2019roberta} to generate text features from each answer segment. Given the RoBERTa's strength in robustly capturing contextual information and semantic nuances in various NLP task~\cite{lee2023learning}, it demonstrated superior performance compared to other embedding methods and large language models~(LLMs) as shown in Section~\ref{sec:encoder}. We extract features from the final layer, focusing on the [CLS] token, resulting in an 85 $\times$ 768-dimensional vector for each answer, where 85 represents the dimensionality of the question embedding.



    \subsection{Question-Aware Module Layer}

        In Figure~\ref{fig:overall_architecture}, a transformer encoder with $h$ multi-heads is utilized to capture attention between questionnaire responses. Initially, a stack of 1-dimensional convolutional layers is applied to process local information, converting varying shapes of HIQ Visual Rep. ($85 \times 275$), HIQ Visual Rep. ($85 \times 88$), and HIQ Visual Rep. ($85 \times 768$) into uniform shapes of $85 \times 4$ denoted as $U_{m}$, $m \in {t, a, v}$.
        
        Subsequently, the question-aware self-attention mechanism guides the transformer encoder to focus on important segments and relationships among the question-embedded sequences. Given that each representation is embedded based on 85 questions, self-attention allows the model to focus on important questions within the question-embedded representation. As a result, this particular attention mechanism enables HiQuE to extract meaningful information, represented as Q-A $M$ Rep; $M \in \left\{Audio, Visual, Text \right\}$, in the form of $85 \times 85$ matrices, for depression detection from each question. We analyze this unimodal attention score to identify the significant components of intra-modality. The same input is employed for self-attention as query ($Q$), key ($K$), and value ($V$) in the following equations:
        \begin{equation}
            \label{eq1}
                MultiHead(Q, K, V) = Concatenate(head_1, ..., head_h)
        \end{equation}
        \begin{equation}
            \label{eq2}
                head_i = Attention(Q W^Q_i, K W^K_i, V W^V_i) 
        \end{equation}
        \begin{equation}
            \label{eq3}
                \text{Q-A} \ m \ Rep. = MultiHead(U_m, U_m, U_m) + U_m 
        \end{equation}

\begin{table*}[t]
    \caption{Performance comparisons on the DAIC-WOZ dataset among twelve state-of-the-art baseline models and the proposed model. (WA*: Weighted Average)} 
    \label{tab:performance}
    \centering
    \tiny
    \setlength{\tabcolsep}{4pt}
    \renewcommand{\arraystretch}{0.95}
    \resizebox{0.95\textwidth}{!}{%
    \begin{tabular}{l||c||c|c|c|c|c|c|c}

    \hline
    \textbf{Method} &\textbf{Approach} & \textbf{Precision} & \textbf{Recall} & \textbf{F1-Score} & \textbf{WA* Prec. ($\uparrow$)} & \textbf{WA* Rec. ($\uparrow$)} & \textbf{WA* F1 ($\uparrow$)} & \textbf{G-Mean ($\uparrow$)} \\ 
    \hline
    
    \textbf{\textit{TFN}}~\cite{zadeh2017tensor} &\textit{Modality-Aware} & 0.67     & 0.73      & 0.68  & 0.84   & 0.78   & 0.81    & 0.699    \\ 
    
    \textbf{\textit{BiLSTM-1DCNN}}~\cite{lin2020towards} &\textit{Modality-Aware} & 0.65      & 0.61      & 0.62     & 0.77   & 0.71 & 0.73  & 0.630 \\ 
    
    \textbf{\textit{MulT}}~\cite{tsai2019multimodal} &\textit{Modality-Aware} & 0.73      & 0.74      & 0.74     & 0.81   & 0.77   & 0.77   & 0.735 \\ 
    
    \textbf{\textit{MISA}}~\cite{hazarika2020misa} &\textit{Modality-Aware} & 0.74      & 0.77      & 0.74    & \textbf{0.86}   & 0.77  & 0.79  & 0.755  \\ 
    
    \textbf{\textit{D-vlog}}~\cite{yoon2022d} &\textit{Modality-Aware} & 0.73      & 0.72      & 0.73     & 0.82   & 0.76 & 0.77  & 0.725   \\
    \hline
    \textbf{\textit{bc-LSTM}}~\cite{poria2017context} &\textit{Context-Aware} & 0.59      & 0.60      & 0.59   & 0.77  & 0.69  & 0.72   & 0.595    \\ 
    
    \textbf{\textit{Emotion Recognition}}~\cite{satt2017efficient} &\textit{Context-Aware} & 0.65      & 0.69      & 0.66   & 0.69   & 0.70  & 0.71  & 0.670    \\ 
    
    \textbf{\textit{Sequence Modeling}}~\cite{al2018detecting} &\textit{Context-Aware} & 0.67  &0.71  &0.70   & 0.85 & 0.73 & 0.77  & 0.690 \\
    
    \textbf{\textit{Topic Modeling}}~\cite{gong2017topic} &\textit{Context-Aware} &0.63   &0.60   &0.62   & 0.81   & 0.71  & 0.74  & 0.615 \\
    
    \textbf{\textit{Context Aware}}~\cite{lam2019context} &\textit{Context-Aware} &0.71   &0.71   &0.71   & 0.85  & 0.73  & 0.77   & 0.710 \\
    
    \textbf{\textit{Speechformer}}~\cite{chen2022speechformer} &\textit{Context-Aware} & 0.70      & 0.72      & 0.70     & 0.78   & 0.76  & 0.76  & 0.710 \\
    
    \textbf{\textit{GRU/BiLSTM-based}}~\cite{shen2022automatic} &\textit{Context-Aware} & 0.75      & 0.78      & 0.75     & \textbf{0.86} & 0.77 & 0.80  & 0.765   \\
    \hline
    \textbf{\textit{HiQuE}} &\textit{Modality + Context} & \textbf{0.78}  & \textbf{0.80}     & \textbf{0.79}  & 0.85  & \textbf{0.80}  & \textbf{0.82}  & \textbf{0.790}\\
    \hline
    \end{tabular}%
}
\end{table*}

    \subsection{Cross-Modal Attention Layer}

        In Figure~\ref{fig:overall_architecture}, the multimodal transformer encoder with $h$ multi-heads integrates information from two modalities using a cross-attention mechanism~\cite{hasan2021humor}. This mechanism allows the model to discern crucial relationships between $m_1$ and $m_2$ modalities, with $m_1$ serving as the source (query) and $m_2$ as the target (key and value).         
        Furthermore, since the information in the two modalities differs, we conduct bidirectional cross-attention between $m_1$ and $m_2$ (i.e., audio-visual, visual-text, and text-audio) to allow the model to learn relevant information across modalities as follows:
        \begin{align}
            \label{eq5}
            U_{m_{1},m_{2}} = MultiHead(U_{m_{1}}, U_{m_{2}}, U_{m_{2}}) + U_{m_{1}}\\
            U_{m_{2},m_{1}} = MultiHead(U_{m_{2}}, U_{m_{1}}, U_{m_{1}}) + U_{m_{2}}
        \end{align}

        Given that the input to the cross-modal attention layer is $U_{m}$, $m \in {t, a, v}$ from the question-aware module layer, each input representation has a shape of $85 \times 85$. This allows us to analyze the multimodal attention score to identify significant components between different modalities.

    \subsection{Depression Detection Layer}
        In the last stage, the $audio$-$visual$, $visual$-$text$, and $text$-$audio$ cross-modal representations are transformed into a final multimodal representation after layer normalization, concatenation, and GAP (global average pooling), as follows:   
        \begin{equation}
            \widetilde{U} = \sum GAP({{U_{m_{1},m_{2}}}\oplus U_{m_{2},m_{1}}})
        \end{equation}
        Finally, multimodal representation is fed into HiQuE's depression detection layer to detect depression as follows:
        \begin{equation}
            \widehat{Y} = softmax(HiQuE(\widetilde{U}))
        \end{equation}

        where the HiQuE prediction layer comprises a fully connected layer and a dropout layer. Since the depression detection task can defined as a binary classification problem, we employed the cross entropy as the loss function as follows:
        \begin{equation}
            \text{Loss} = -\frac{1}{b} \sum_{i=1}^{b} \left[ y_i \cdot \log(\hat{y}_i) + (1 - y_i) \cdot \log(1 - \hat{y}_i) \right]
        \end{equation} 

        where $b$ represents the batch size, $i$ is an index representing each example within the batch, $y_i$ is the actual label where 0 represents normal and 1 represents depression, and $\hat{y_i}$ is the softmax function that represents the model's prediction or probability.

\section{Experiments}

        We use Tensorflow~\cite{abadi2016tensorflow} to implement the proposed model. The dropout rate, batch size, epochs, and learning rate were set to 0.5, 8, 100, and 0.0002, respectively. The maximum sequence length was set to 85 since all sequences are embedded into 85 questions. All weights are randomly initialized in both \textit{HiQuE} and baselines.

    \subsection{Baseline Methods}
        
        To evaluate the overall performance of the proposed model, we compare its performance against five state-of-the-art multimodal models for depression detection and emotion recognition as follows:  (i) Tensor Fusion Network (\textbf{\textit{TFN}})~\cite{zadeh2017tensor}, (ii) bidirectional LSTM / 1D CNN-based model (\textbf{\textit{BiLSTM-1DCNN}})~\cite{lin2020towards}, (iii) Multimodal Transformer (\textbf{\textit{MulT}})~\cite{tsai2019multimodal}, (iv) \textbf{\textit{MISA}}~\cite{hazarika2020misa}, and (v) \textbf{\textit{D-vlog}}~\cite{yoon2022d}. Since these models were specifically designed to analyze multimodal fusion methods, we have categorized them as \textbf{``\textit{Modality-Aware}''}. 

        We further utilize seven context-aware multimodal models for depression detection and emotion recognition to compare the analysis of the hierarchical structure of clinical interviews: (i) bidirectional contextual LSTM (\textbf{\textit{bc-LSTM}})~\cite{poria2017context}, (ii) \textbf{\textit{Emotion Recognition}}~\cite{satt2017efficient}, (iii) \textit{\textbf{Sequence Modeling}}~\cite{al2018detecting}, (iv) \textit{\textbf{Topic Modeling}}~\cite{gong2017topic}, (v) Context-aware deep learning (\textit{\textbf{Context-Aware}})~\cite{lam2019context}, (vi) \textbf{\textit{Speechformer}}~\cite{chen2022speechformer} and (vii) \textbf{\textit{GRU/BiLSTM-based}}~\cite{shen2022automatic}. As these methods consider the context of interview questions and answers or focus on the topics of questions and the timing of their appearance during the interview, we have categorized them as \textbf{``\textit{Context-Aware}''}. 

        Note that we extract multimodal features from the entire interview sequence to train the five modality-aware methods and seven context-aware methods. All models were trained on the same data partition to ensure fairness and evaluated using the hyper-parameters that showed the best performance.

    \begin{table}[t]
        \caption{Performance on different text embeddings.}
        \label{tab:encoder}
        \centering
        \resizebox{7cm}{!}{
        \begin{tabular}{c|c|c|c}
        \toprule
        \begin{tabular}[c]{@{}c@{}}\textbf{Text Embedding}
        \end{tabular} & \textbf{Precision} & \textbf{Recall} & \textbf{F1-Score} \\ \midrule
            BART~\cite{lewis2019bart}  & 0.75   & 0.76  & 0.74   \\  
            
            GloVe~\cite{pennington2014glove} & 0.72  & 0.73 & 0.72     \\
            
            BERT~\cite{devlin2018bert} & 0.76   & 0.77 & 0.76  \\
            \midrule
            
            CodeLlama~\cite{roziere2023code} & \textbf{0.78}   & 0.75    & 0.76     \\
            
            Llama2~\cite{touvron2023llama} & 0.77   & 0.77    & 0.77     \\
            
            GPT-2~\cite{radford2019language} & 0.76   & 0.78    & 0.77     \\
            \midrule
            
            \textbf{RoBERTa~\cite{liu2019roberta} (Ours)} & \textbf{0.78}  & \textbf{0.80} & \textbf{0.79}  \\
        \bottomrule
        \end{tabular}
        }
    \end{table}

\begin{table*}[t]
\caption{Validating the generalizability of \textit{HiQuE} on the two datasets (E-DAIC-WOZ and MIT Interview) using multimodal inputs (audio, video, and text). Tasks include binary classification for Depression Detection, Stress Level Prediction, and Job Interview Performance, respectively.}

\label{tab:edaicwoz}
\centering
\tiny
\setlength{\tabcolsep}{4pt}
\renewcommand{\arraystretch}{0.95}
\resizebox{0.8\textwidth}{!}{

\begin{tabular}{c||c|c|c|c|c|c|c|c|c}

\hline
\multirow{4}{*}{\textbf{Methods}} & \multicolumn{3}{c|}{\textbf{E-DAIC-WOZ~\cite{ringeval2019avec}}}             & \multicolumn{6}{c}{\textbf{MIT Interview dataset~\cite{naim2016automated}}}                                                                  \\ \cline{2-10} 
                         & \multicolumn{3}{c|}{\textbf{Depression Detection}} & \multicolumn{3}{c|}{\textbf{Stress Level Prediction}} & \multicolumn{3}{c}{\textbf{\begin{tabular}[c]{@{}c@{}}Job Interview\\ Performance Prediction\end{tabular}}} \\ \cline{2-10} 
                         & \multicolumn{3}{c|}{Overall}              & \multicolumn{3}{c|}{Overall}                 & \multicolumn{3}{c}{Overall}                    \\ \cline{2-10} 
                         & \textit{Pre.} & \textit{Rec.} & \textit{F1.} & \textit{Pre.} & \textit{Rec.} & \textit{F1.} & \textit{Pre.} & \textit{Rec.}   & \textit{F1.}    \\ \hline
\textbf{\textit{GRU/BiLSTM-based}}~\cite{shen2022automatic}  &0.67  &0.63  &0.65   &0.70  &0.76  & 0.73   &0.73  &0.75  & 0.74   \\ 
\textbf{\textit{D-vlog}}~\cite{yoon2022d}                    &0.65  &0.69  &0.67   &0.71  &0.73  & 0.72   &0.71  &0.72  & 0.72   \\ 
\textbf{\textit{MISA}}~\cite{hazarika2020misa}               &0.62  &0.64  &0.63   &0.69  &0.68  & 0.68   &0.69  &0.69  & 0.69   \\
\textbf{\textit{MulT}}~\cite{hazarika2020misa}               &0.64  &0.65  &0.64   &0.70  &0.73  & 0.72   &0.70  &0.72  & 0.71   \\ \hline
\textbf{\textit{HiQuE}}                                      &\textbf{0.71}  &\textbf{0.70}  &\textbf{0.70}  &\textbf{0.75}  &\textbf{0.81}  &\textbf{0.78} &\textbf{0.76}  &\textbf{0.79}  &\textbf{0.77} \\ \hline
\end{tabular}
}
\end{table*}

    \subsection{Experimental Results}
            To provide a comprehensive assessment of the models' performance, particularly in the context of an imbalanced dataset (i.e., DAIC-WOZ), we report experimental results with various metrics including the weighted average and geometric mean scores (G-mean score). Table~\ref{tab:performance} shows the Macro Average precision / recall / F1-score,  Weighted Average precision / recall / F1-score, and G-mean score of the baseline models and the proposed model, respectively.

            As shown in Table~\ref{tab:performance}, \textbf{\textit{HiQuE}} achieves the best depression detection with a macro average F1-score of 0.79, a weighted average F1-score of 0.82, and a G-mean score of 0.790. As macro-average treats each class equally, while weighted-average gives weight based on class size, the result that \textit{HiQuE} excels in both metrics highlights \textit{HiQuE}'s robustness and effectiveness against an imbalanced dataset. showcasing its ability to capture distinct depression indicators.

            Among the baseline models, \textbf{\textit{GRU/BiLSTM-based}}~\cite{shen2022automatic} achieves the highest performance with a macro average F1-score of 0.75, weighted average precision of 0.86, and G-mean score of 0.765. This underscores the effectiveness of analyzing speech characteristics and linguistic content within individual utterances for depression diagnosis within interview sequences. Furthermore, \textbf{\textit{MISA}} demonstrates the second-highest performance among the baselines with a macro average F1-score of 0.74, weighted average precision of 0.86, and G-mean score of 0.755. This suggests that incorporating two subspaces (modality-invariant and modality-specific) allows the model for a comprehensive understanding of multimodal data, which suggests that for accurate multimodal depression detection, it is essential to capture both inter- and intra-representations between modalities. 
            We also find that \textbf{\textit{MulT}} and \textbf{\textit{D-vlog}} exhibit promising performance at 0.74 and 0.73 of the macro average F1-score, respectively. This suggests that employing a cross-attention mechanism to learn the relationship between multiple modalities helps the model learn important signals for depression detection.

        \subsection{Text Embedding Performance Comparison}
        \label{sec:encoder}
    
        We chose to use the pre-trained RoBERTa~\cite{liu2019roberta} as an encoder due to its higher performance as shown in in Table~\ref{tab:encoder}, in comparison with other popular embedding techniques and large language models (LLMs). The high performance of RoBERTa is due to its robust representations and comprehensive contextual understanding. LLMs also showed a comparable performance as shown in Table~\ref{tab:encoder}, but we decided not to use them due to practical challenges related to privacy and stability, particularly in mental health applications.

            
            

            

        \begin{table*}[t]
            \caption{The results of the ablation study on hierarchical question embedding process and model layers as illustrated in Figure~\ref{fig:overall_architecture}, along with augmentation methods with random sampling as described in Section~\ref{augmentation}. The term ``\textit{Q-A Module}'' denotes the Question-Aware Module Layer, while ``\textit{C-M Attention}'' represents the Cross-Modal Attention Layer, which is examined as part of the model components. ``\textit{Q.E.}'' and ``\textit{H.Q.E.}'' stand for Question Embedding and Hierarchical Question Embedding, respectively, and ``\textit{Aug.}'' refers to the Augmentation.}
            \label{tab:ablation}
            \setlength{\tabcolsep}{4pt}
            \renewcommand{\arraystretch}{0.95}
            \fontsize{4pt}{5pt}\selectfont
            \centering
            \resizebox{0.8\textwidth}{!}{
            \begin{tabular}{ccccc||c|c|c|c}
            \hline
            \multicolumn{5}{c||}{Ablation Settings} &
              \multirow{3}{*}{Precision} &
              \multirow{3}{*}{Recall} &
              \multirow{3}{*}{F1-Score} &
              \multirow{3}{*}{WA* F1 ($\uparrow$)} \\ \cline{1-5}
            \multicolumn{2}{c|}{Question Embedding}          & \multicolumn{2}{c|}{Model Layer}              & \multirow{2}{*}{Aug.} &  &  &  &  \\ \cline{1-4}
            \multicolumn{1}{c|}{Q.E.} &
              \multicolumn{1}{c|}{H.Q.E.} &
              \multicolumn{1}{c|}{Q-A Module} &
              \multicolumn{1}{c|}{C-M Attention} &
               &
               &
               &
               &
               \\ \hline
            \multicolumn{1}{c|}{\xmark} & \multicolumn{1}{c|}{\xmark} & \multicolumn{1}{|c|}{\cmark} & \multicolumn{1}{c|}{\cmark} & \cmark &0.74  &0.73  &0.73  &0.75  \\
            
            \multicolumn{1}{c|}{\cmark} & \multicolumn{1}{c|}{\xmark} & \multicolumn{1}{|c|}{\cmark} & \multicolumn{1}{c|}{\cmark} & \cmark &0.75  &0.76  &0.75  &0.77  \\
            

            \multicolumn{1}{c|}{\cmark} & \multicolumn{1}{c|}{\cmark} & \multicolumn{1}{|c|}{\xmark} & \multicolumn{1}{c|}{\xmark} & \cmark &0.73  &0.72  &0.72  &0.74  \\
            
            \multicolumn{1}{c|}{\cmark} & \multicolumn{1}{c|}{\cmark} & \multicolumn{1}{|c|}{\cmark} & \multicolumn{1}{c|}{\xmark} & \cmark &0.73  &0.75  &0.74  &0.77  \\

            \multicolumn{1}{c|}{\cmark} & \multicolumn{1}{c|}{\cmark} & \multicolumn{1}{|c|}{\xmark} & \multicolumn{1}{c|}{\cmark} & \cmark &0.75  &0.76  &0.76  &0.79  \\
            
            \multicolumn{1}{c|}{\cmark} & \multicolumn{1}{c|}{\cmark} & \multicolumn{1}{|c|}{\cmark} & \multicolumn{1}{c|}{\cmark} & \xmark &0.76  &0.77  &0.76  &0.80  \\ \hline

            \multicolumn{1}{c|}{\cmark} & \multicolumn{1}{c|}{\cmark} & \multicolumn{1}{|c|}{\cmark} & \multicolumn{1}{c|}{\cmark} & \cmark &\textbf{0.78}  &\textbf{0.80}  &\textbf{0.79}  &\textbf{0.82}  \\ \hline
   
            \end{tabular}
            }
        \end{table*}

    \subsection{Generalization to Unseen Questions}
    \label{sec:unseen}
    To assess \emph{HiQuE}'s generalizability, we further utilized the E-DAIC-WOZ~\cite{ringeval2019avec} and MIT Interview dataset~\cite{naim2016automated}. The E-DAIC-WOZ~\cite{ringeval2019avec} comprises audio-visual recordings of semi-clinical interviews conducted in English, featuring numerous questions absent in the DAIC-WOZ dataset. However, it does not provide the transcript of interviewer's questions, making it difficult to determine the specific questions asked. The MIT Interview dataset~\cite{naim2016automated} includes 138 interview videos of internship-seeking students from MIT, featuring facial expressions, language use, and prosodic cues. Moreover, it provides ground truth labels for stress level and job interview performance, rated by nine independent judges. This dataset encompasses multimodal features influencing mental states during job interviews~\cite{naim2016automated}.

    We adapt our model to these datasets by extracting text from the audio using the whisper~\cite{radford2023robust} and mapping unseen questions to the predefined list (Table~\ref{tab:questionlist}) based on the BERT-score~\footnote{https://huggingface.co/spaces/evaluate-metric/bertscore} similarity. If a question's similarity falls below the average, we consider it as a new question and add it to the list. Baselines adopted the same encoder as \textit{HiQuE} but without hierarchical question embedding. Table~\ref{tab:edaicwoz} shows that \textit{HiQuE} outperforms the baselines across both datasets for the three different tasks: Depression Detection, Stress Level Prediction, and Job Interview Performance Prediction. This underscores \textit{HiQuE}'s effectiveness in detecting depression cues from clinical interviews, even with non-predefined questions. More importantly, the experimental result that \textit{HiQuE} has shown promising performance not only in clinical interviews but also in job interviews highlights its usability in various real-world interview scenarios.

\begin{figure}[t]
  \centering
  \includegraphics[width=\linewidth]{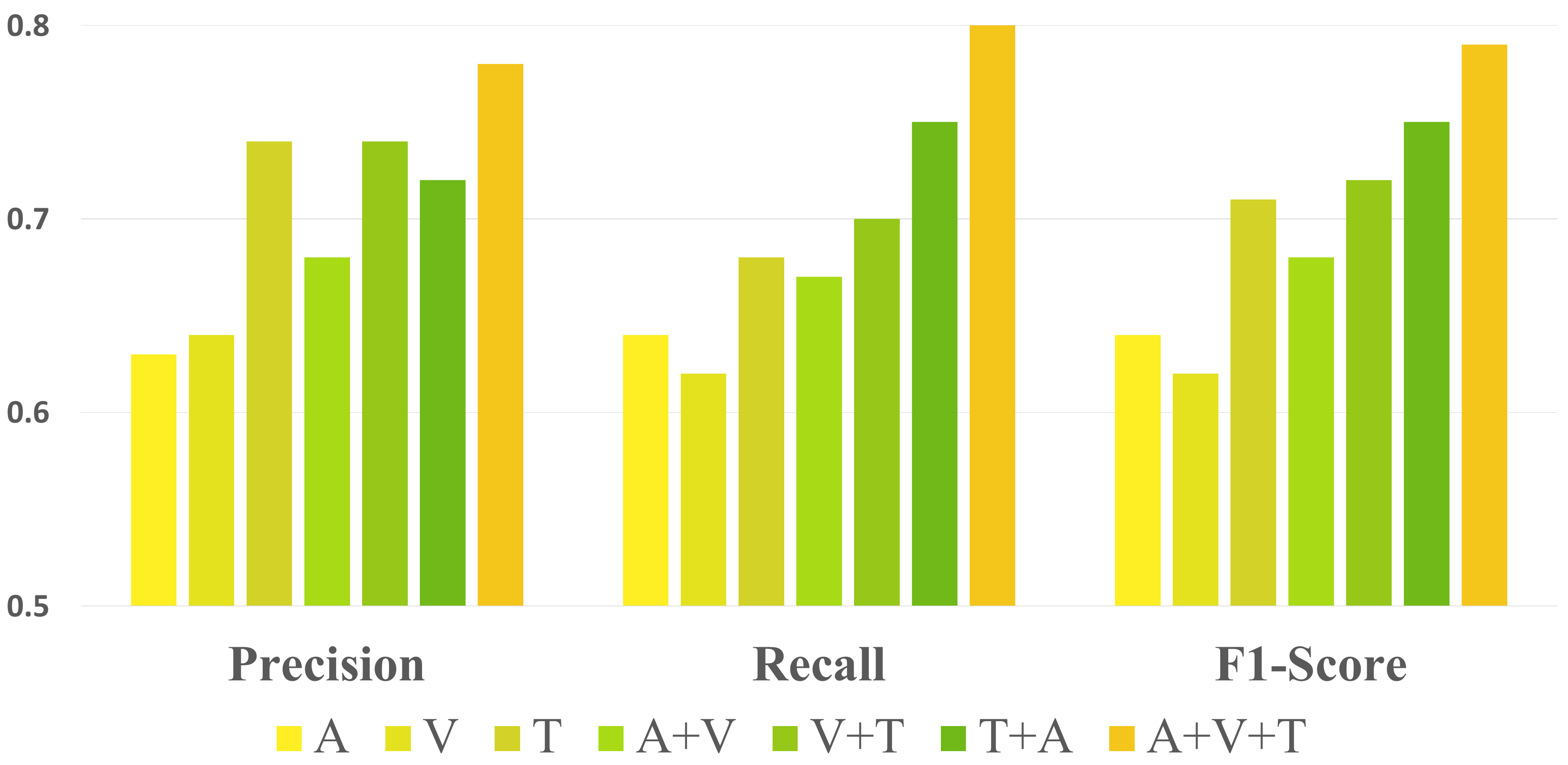}
  \caption{Performance comparisons between unimodal and multimodal depression detection models. A, V, and T denote audio, visual, and text modality, respectively. The X-axis indicates the macro average precision, recall, and F1 score, while the Y-axis represents the corresponding scores.}
  \label{fig:modality}
  \vspace{-0.1in}
\end{figure}

        \begin{figure}[t]
          \centering
          \includegraphics[width=\linewidth]{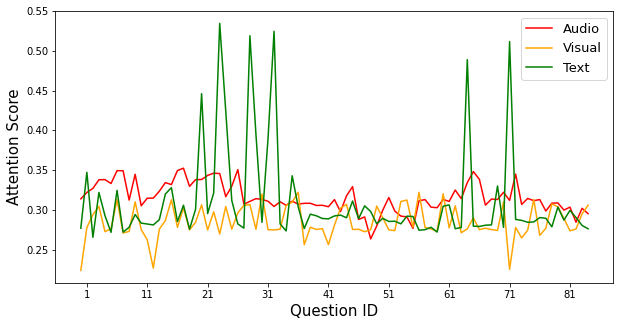}
          \caption{Distributions of attention scores across different modalities in each question, with the X-axis representing the Question ID from the first question ($Q_1$) to the last question ($Q_{85}$), and the Y-axis indicating the attention score.}
          \label{fig:distribution}
          \vspace{-0.1in}
        \end{figure}

\section{Analysis}

        \subsection{Analysis on Different Modalities}
            To analyze the importance of each modality (i.e., audio, visual, and text) for detecting depression, we compare the performance of models that are trained with different sets of modalities. For the \textbf{unimodal models} (i.e., A, V, T), we first simply utilize a hierarchical question embedding process followed by a question-aware module layer for each input modality. We then add global average pooling and fully connected layers with softmax activation function to generate predicted labels (i.e., depressed or not). As shown in Figure~\ref{fig:modality}, the model trained with text achieves the highest performance (0.71 of macro average F1-score) among the unimodal models. This implies that the text modality contains the most useful information in depression detection, which can be linked to the results of the prior studies~\cite{yang2020mtag,rohanian2019detecting,dham2017depression,chiong2021textual}. For \textbf{bimodal models} (i.e., A+V, V+T, A+T), we first fuse two unimodal encoders via a cross-modal attention layer. We then add the same depression detection layer as unimodal models. Since the text feature contains the most useful information, we find that the bimodal models trained with text modality (i.e., A+T, V+T) show higher performance than the model trained without text modality (i.e., A+V). Also, we find that considering \textbf{all modalities} (i.e., A+V+T) significantly improves performance. This reveals that learning both verbal and non-verbal signals, as well as their relationships, is an effective way for depression detection.

   \begin{figure*}[t!]
          \centering
          \includegraphics[width=\linewidth]{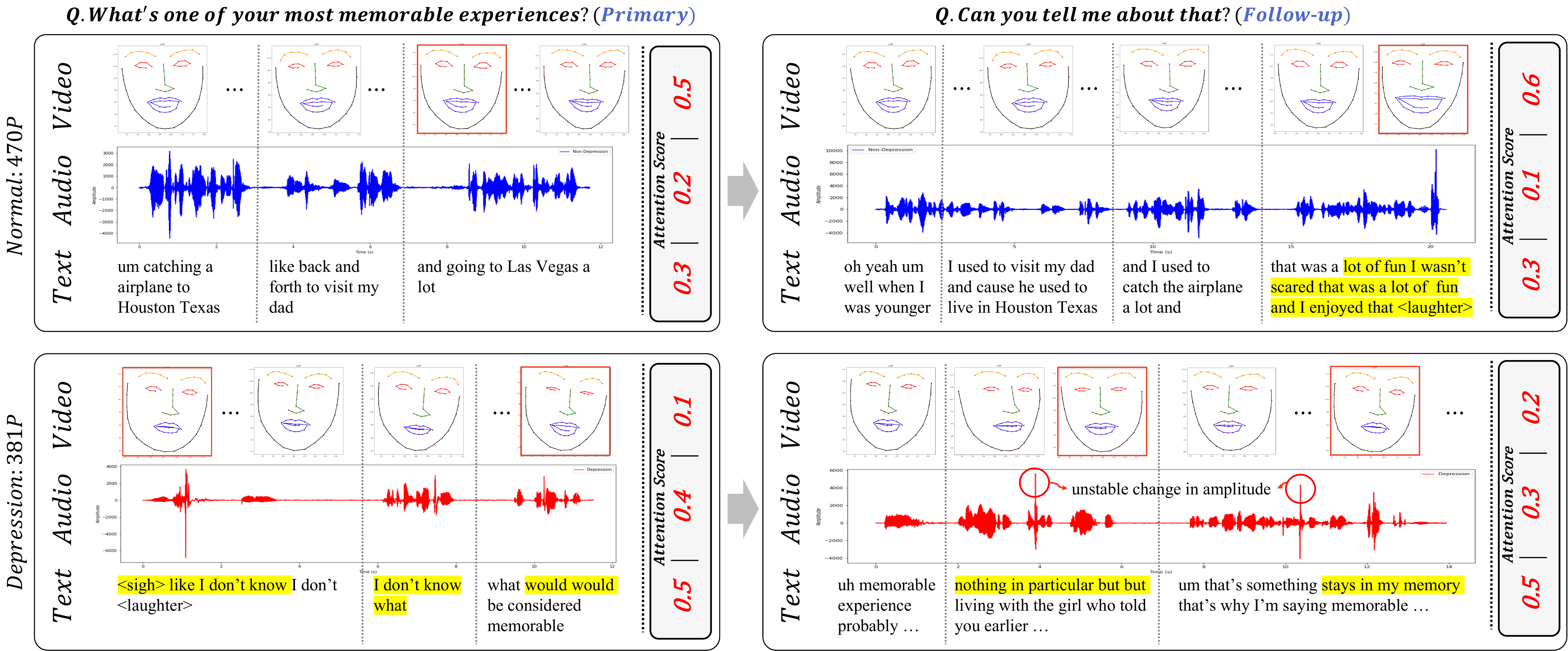}
          \caption{A case analysis of a depression case (381P) and a non-depression case (470P). Specific audio, video, and text responses are provided in response to the primary question, ``\textit{What's one of your most memorable experiences?}'' and the follow-up question, ``\textit{Can you tell me about that?}''. The highlighted sections in the figures (i.e., red rectangles in facial features or yellow highlights in the textual features) indicate distinct characteristics between a depression case and a non-depression case. 
          }
          \label{fig:case-specific}
        \end{figure*}

\subsection{Intermodal Interaction Analysis}

            By examining the attention score distributions across different modalities, as depicted in Figure~\ref{fig:distribution}, we highlight the significance of each modality in depression detection. Notably, questions directly related to emotions or past experiences, such as \textit{``Tell me about an event or something that you wish you could erase from your memory?''} or \textit{``Tell me about the last time you felt really happy?''}, had a significant impact on both audio and visual modalities, while those related to a current emotional state or past depression diagnosis, such as \textit{``Have you been diagnosed with depression?''} or \textit{``How have you been feeling lately?''}, had the highest impact on text.
            
            We also explore the impact of individual modalities (i.e., audio, visual, and text) when the model fails to make accurate predictions. Depressed patients are often misclassified as normal when interviewees exhibit cheerful tones or frequent laughter (resulting in high audio and visual attention scores, Figure~\ref{fig:distribution}), or when there are no clear indicators of depression during the interview. Conversely, our model tends to misclassify normal as depression when negative words are frequently used, particularly when participants express recent feelings of anxiety and depression. In these cases, as highlighted in Figure~\ref{fig:distribution}, the text attention score predominantly influences the incorrect predictions.

        \subsection{Ablation Study}
        \subsubsection{Hierarchical Question Embedding}
             To highlight the benefits of our proposed hierarchical question embedding process, we conducted an ablation study with three distinct cases: \textit{Non-Question Embedding} (N.Q.E), only \textit{Question Embedding} (Q.E.), and \textit{Hierarchical Question Embedding} (H.Q.E.), as shown in Table~\ref{tab:ablation}. In the case of \textbf{\textit{N.Q.E}}, the entire interview sequence is treated as a single sequence for the depression detection model. Specifically, in this case, the input sequences are cropped from the beginning of the utterance to the end of the interview conversation. When only \textbf{\textit{Q.E}} is applied, the interview is segmented into question-answer pairs. Notably, this procedure only divides the sequence into questions and aligns them with the respective question topics without incorporating hierarchical positional embedding. The improvement presented in Table~\ref{tab:ablation} highlights that, for depression detection with clinical interviews, extracting useful information based on a question-driven approach is more effective than considering the entire interview sequence as a single sequence. In the last scenario \textbf{\textit{H.Q.E}}, hierarchical position embedding is introduced following the question embedding procedure. To elaborate, after dividing the interview into question-answer pairs using question embedding, a hierarchical relationship (primary or follow-up) among the questions is tagged through hierarchical position embedding. As shown in Table~\ref{tab:ablation}, our proposed hierarchical question embedding process effectively forces the model to capture hierarchical relationships and the importance of the questions.

        \subsubsection{Model Components}
             To assess the effectiveness of each layer in the HiQuE, we conducted an ablation study on the \textit{Question-Aware Module Layer} and the \textit{Cross-Modal Attention Layer}. As shown in Table~\ref{tab:ablation}, without \textbf{``\textit{Q-A Module Layer}''}, each audio, visual, and text representation undergoes hierarchical question embedding and feature extraction processes before entering the Cross-Modal Attention Layer, which incorporates bidirectional cross-attention (e.g., $U_{a,v}\ \text{-} \ U_{v,a}$) as illustrated in Figure~\ref{fig:overall_architecture}. Given that the \textit{Question-Aware Module Layer} assesses the relevance, importance, and mutual influence of the 85 embedded questions for each modality, its absence results in a performance degradation of the model. Without \textbf{``\textit{C-M Attention Layer}''}, the HIQ Audio Rep., HIQ Visual Rep., and HIQ Text Rep. go through the \textit{Question-Aware Module Layer}, and are then concatenated before entering the \textit{Depression Detection Layer}. Since the \textit{Cross-Modal Attention Layer} computes relevant information from different modalities, the result highlights the effectiveness of considering information from both different modalities for accurate depression detection. Interestingly, as depicted in Table~\ref{tab:ablation}, the setting without \textbf{``\textit{Q-A Module Layer}''} achieves a higher macro average F1-score and weighted average F1-score compared to the setting without \textbf{``\textit{C-M Attention Layer}''}. This reveals that in detecting depression, it is more important to learn relevant information and interactions between modalities than to analyze the relationships and importance of each question.
             Furthermore, Table~\ref{tab:ablation} also presents the performances of our proposed model with and without data augmentation. The results confirm that data augmentation enhances performance by balancing the sizes of depression and non-depression cases in the training set.

\section{Case Study}
\label{sec:casestudy}

In this section, we present a case study on samples from our test set to assess the effectiveness of the decision-making process of \textit{HiQuE}. Specifically, we examine the verbal (i.e., text) and non-verbal (i.e., audio and video) signals for the two cases: a depressed individual (381P) and a non-depressed individual (470P). Our analysis focuses on the distinct attributes of audio, text, and visual attention scores for each individual. For a fair comparison, we apply a normalization technique to the amplitude and time of the audio waves, allowing for unbiased and consistent analysis and comparison. 

Figure~\ref{fig:case-specific} showcases how the model integrates text, audio, and visual features during the decision-making process for each questionnaire response. In the case of the primary question ``\textit{What's one of your most memorable experiences?}'', we observe that the depressed individual faces difficulties in providing a prompt response. He/she exhibits hesitation while reflecting on memorable experiences and ultimately struggles to provide a specific answer. In contrast, the non-depressed individual is more likely to respond immediately and accurately.

By analyzing audio, visual, and text attention scores to the follow-up question ``\textit{Can you tell me about that?}'', we observe the comprehensive exploration and understanding of various responses exhibited by \textit{HiQuE} in detecting depression. In the case of a non-depressed individual (470P), detailed explanations, expressions of excitement, and smiling faces are evident in the answer to the follow-up question. Note that \textit{HiQuE} also gives the highest attention score 0.6 to visual features. Furthermore, apart from \textit{<laughter>}, the audio waves display symmetrical patterns without irregular fluctuations, indicating a more wide range of tones and amplitudes. On the other hand, the depressed individual (381P) encounters difficulties recalling memorable experiences when responding to the follow-up question. Instead of positive recollections, this individual shares memories of regrettable past incidents. By examining the audio wave of the depressed individual, we observe unstable fluctuations in amplitude while his/her facial expressions remain neutral. For this reason, \textit{HiQuE} assigns the highest attention score of 0.5 to text features, followed by attention scores of 0.3--0.4 for audio features.

Our analysis of the attention scores for each modality during the model's diagnostic process demonstrates that \textit{HiQuE} effectively incorporates the interview structure through its hierarchical question embedding layer. The case study provides further evidence that \textit{HiQuE} successfully captures the sequential information of all questions and maximizes the mutual information between modalities by leveraging the question-aware module and the cross-modal attention layer.


\section{Conclusion}
In this paper, we presented HiQuE, a novel hierarchical question embedding model for multimodal depression detection. HiQuE efficiently captures the hierarchical structure of questions in clinical interviews and explores the correlations between different modalities to extract valuable information for depression detection. 
Through a comprehensive case study, we confirmed that the HiQuE focuses on questions specifically related to depression and makes its final decision by utilizing attention scores. This approach allows the model to mimic the expertise of clinical professionals during clinical interviews, where the interaction of questionnaire responses plays a crucial role. 
Given HiQuE's demonstrated generalizability to unseen questions, future plans involve extending its applicability to additional speech-related tasks and exploring the advantages of hierarchical question embedding further.

\begin{acks}


This research was supported by the MSIT (Ministry of Science, ICT), Korea, under the Global Research Support Program in the Digital Field program (RS-2024-00425354) and the Graduate School of Metaverse Convergence support program (IITP-2024-RS-2023-00254129) supervised by the IITP (Institute for Information \& Communications Technology Planning \& Evaluation).
\end{acks}


\appendix
\section{List of the 85 Questions in DAIC-Woz Dataset}
\label{appendix:f}
We present the comprehensive list of 85 questions employed in the DAIC-WOZ dataset. Each question posed by the interviewer (Ellie) was mapped to the relevant topics using the question's topic codes~\cite{gong2017topic}. We augmented the list with additional questions from the interviewer (Ellie) and rectified any inaccuracies in the existing questions. Furthermore, we organized all the questions into a hierarchical structure, comprising 66 primary questions and 19 follow-up questions, determined by their content and the order in which they were posed.

\begin{table}[h]
    \tiny
    \caption{Question list in the DAIC-Woz dataset. All questions were categorized into 66 main questions and 19 follow-up questions using hierarchical question mapping.}
    \centering
    \label{tab:questionlist}
    \resizebox{0.48\textwidth}{!}{%
    \begin{tabular}{clc}
    \toprule
    Ind.   & Question   &Type        \\ 
    \midrule
    (1) & how has seeing a therapist affected you     & Primary     \\                     
    (2) & tell me about the last time you felt really happy     & Primary    \\            
    (3) & where are you from originally    & Primary                                     \\           
    (4) & when was the last time you argued with someone and what was it about     & Primary     \\    
    (5) & what advice would you give to yourself ten or twenty years ago      & Primary \\        
    (6) & how are you at controlling your temper       & Primary               \\             
    (7) & what are some things you really like about l\_a      & Primary        \\           
    (8) & what are you most proud of in your life     & Primary                            \\         
    (9) & who’s someone that’s been a positive influence in your life    & Primary\\    
    (10) & how would your best friend describe you     & Primary                  \\  
    (11) & what are some things you don’t really like about l\_a     & Primary      \\
    (12) & what did you study at school       & Primary                                                 \\
    (13) & is there anything you regret  & Primary                                                \\
    (14) & what’s your dream job      & Primary                                                  \\
    (15) & what do you enjoy about traveling        & Primary                                  \\
    (16) & what are you like when you don’t sleep well      & Primary                 \\
    (17) & what’s one of your most memorable experiences       & Primary                        \\
    (18) & tell me about the hardest decision you’ve ever had to make       & Primary     \\
    (19) & what are some things you like to do for fun   & Primary                           \\
    (20) & tell me about a situation that you wish you had handled differently      & Primary\\  
    (21) & tell me about an event or something that you wish you could erase from your memory     & Primary\\  
    (22) & why did you move to l\_a    & Primary                                            \\
    (23) & what are some things you wish you could change about yourself      & Primary       \\
    (24) & what would you say are some of your best qualities      & Primary                  \\
    (25) & how often do you go back to your home town       & Primary                           \\
    (26) & how long ago were you diagnosed      & Primary                              \\
    (27) & what’s something you feel guilty about     & Primary                                   \\
    (28) & when did you move to l\_a        & Primary                                          \\
    (29) & how easy was it for you to get used to living in l\_a      & Primary               \\
    (30) & when was the last time you felt really happy        & Primary                   \\
    (31) & what’s the hardest thing about being a parent       & Primary                       \\
    (32) & do you still go to therapy now       & Primary                                     \\
    (33) & do you travel a lot     & Primary                                               \\
    (34) & have you ever served in the military     & Primary                       \\
    (35) & when was the last time that happened       & Primary                             \\
    (36) & what’s the best thing about being a parent      & Primary                           \\
    (37) & what are some things that make you really mad    & Primary                                \\
    (38) & do you find it easy to be a parent       & Primary                                  \\
    (39) & what do you do now     & Primary                                                            \\
    (40) & what were your symptoms     & Primary                                                   \\
    (41) & tell me how you spend your ideal weekend        & Primary                          \\
    (42) & what do you do when you are annoyed      & Primary                                   \\
    (43) & tell me about your kids   & Primary \\
    (44) & tell me about a time when someone made you feel really badly about yourself  &Primary \\
    (45) & what are some ways that you’re different as a parent than your parents   &Primary\\
    (46) & what do you think of today’s kids  &Primary\\
    (47) & do you feel down  &Primary\\
    (48) & how do you like your living situation   &Primary\\
    (49) & how are you doing today   &Primary\\
    (50) & do you have roommates   &Primary\\
    (51) & do you think that maybe you’re being a little hard on yourself  &Primary\\
    (52) & do you have disturbing thoughts   &Primary\\
    (53) & where do you live  &Primary\\
    (54) & what did you do after the military   &Primary\\
    (55) & did you ever see combat  &Primary\\
    (56) & why don’t we talk about that later   &Primary\\
    (57) & how did serving in the military change you   &Primary\\
    (58) & have you noticed any changes in your behavior or thoughts lately   &Primary\\
    (59) & have you been diagnosed with depression  &Primary\\
    (60) & how easy is it for you to get a good night sleep   &Primary\\
    (61) & how close are you to your family   &Primary\\
    (62) & how have you been feeling lately   &Primary\\
    (63) & do you consider yourself an introvert &Primary\\
    (64) & have you ever been diagnosed with p\_t\_s\_d    &Primary\\
    (65) & do you feel like therapy is useful   &Primary\\
    (66) & what do you do to relax &Primary\\
    (67) & can you tell me about that   &Follow-up\\
    (68) & why   &Follow-up\\
    (69) & how hard is that    &Follow-up\\
    (70) & what made you decide to do that   &Follow-up\\
    (71) & are you still doing that  &Follow-up\\
    (72) & what got you to seek help   &Follow-up \\
    (73) & how do you cope with them   &Follow-up\\
    (74) & how does it compare to l\_a   &Follow-up\\
    (75) & are you okay with this &Follow-up\\
    (76) & are they triggered by something  &Follow-up\\
    (77) & are you happy you did that   &Follow-up\\
    (78) & could you have done anything to avoid it   &Follow-up\\
    (79) & has that gotten you in trouble    &Follow-up\\
    (80) & how do you know them   &Follow-up\\
    (81) & do you feel that way often  &Follow-up\\
    (82) & did you think you had a problem before you found out &Follow-up \\
    (83) & why did you stop &Follow-up\\
    (84) & what’s it like for you living with them   &Follow-up \\
    (85) & can you give me an example of that &Follow-up \\

    \bottomrule
    \end{tabular}%
    }
    \vspace{0.05pt}
    \end{table}

\bibliographystyle{ACM-Reference-Format}
\bibliography{0_main}

\end{document}